\documentclass[letterpaper]{article} 
\usepackage{aaai2026}  
\usepackage{times}  
\usepackage{helvet}  
\usepackage{courier}  
\usepackage[hyphens]{url}  
\usepackage{graphicx} 
\usepackage{natbib}  
\usepackage{caption} 
\usepackage{algorithm}
\usepackage{algorithmic}
\usepackage{amsmath}
\usepackage{bm}
\usepackage{multirow} 
\usepackage{booktabs}
\usepackage{amssymb}
\usepackage[skins]{tcolorbox}
\def\model{BEAT}
\usepackage{newfloat}
\usepackage{listings}
\DeclareCaptionStyle{ruled}{labelfont=normalfont,labelsep=colon,strut=off} 
\floatstyle{ruled}
\newfloat{listing}{tb}{lst}{}
\floatname{listing}{Listing}
\title{Behavior Tokens Speak Louder: Disentangled Explainable Recommendation with Behavior Vocabulary}
\author{
    Xinshun Feng\textsuperscript{\rm 1},
    Mingzhe Liu\textsuperscript{\rm 1}\thanks{Corresponding Authors: Mingzhe Liu},
    Yi Qiao\textsuperscript{\rm 2},
    Tongyu Zhu\textsuperscript{\rm 2},
    Leilei Sun\textsuperscript{\rm 2},
    Shuai Wang\textsuperscript{\rm 1}
}
\affiliations{
    \textsuperscript{\rm 1}Hangzhou International Innovation Institute, Beihang University, Hangzhou, China\\
    \textsuperscript{\rm 2}State Key Laboratory of Complex \& Critical Software Environment, Beihang University, Beijing, China\\
    \{xinshunfeng, mzliu1997, qiaoy, zhutongyu, leileisun, shuaiwang\}@buaa.edu.cn
}

\usepackage{bibentry}

\begin{document}

\maketitle

\begin{abstract}
Recent advances in explainable recommendation have explored the integration of language models to analyze natural language rationales for user–item interactions.
Despite their potential, existing methods often rely on ID-based representations that obscure semantic meaning and impose structural constraints on language models, thereby limiting their applicability in open-ended scenarios.
These challenges are intensified by the complex nature of real-world interactions, where diverse user intents are entangled and collaborative signals rarely align with linguistic semantics.
To overcome these limitations, we propose \model, a unified and transferable framework that tokenizes user and item behaviors into discrete, interpretable sequences. 
We construct a behavior vocabulary via a vector-quantized autoencoding process that disentangles macro-level interests and micro-level intentions from graph-based representations. 
We then introduce multi-level semantic supervision to bridge the gap between behavioral signals and language space.
A semantic alignment regularization mechanism is designed to embed behavior tokens directly into the input space of frozen language models.
Experiments on three public datasets show that \model~improves zero-shot recommendation performance while generating coherent and informative explanations.
Further analysis demonstrates that our behavior tokens capture fine-grained semantics and offer a plug-and-play interface for integrating complex behavior patterns into large language models.
\end{abstract}

\begin{links}
    \link{Code}{https://github.com/fxsxjtu/BEAT}
\end{links}

\section{Introduction}

With the rapid growth of online content, users are confronted with information overload, which significantly hinders their ability to identify the most relevant items. 
Recommendation systems have been widely adopted to help users filter out irrelevant products from vast choices~\cite{2015recommendation, 2019recommendation}. 
Various approaches have been explored to enhance recommendation accuracy by modeling users' latent interests and preferences, leading to a satisfying user experience~\cite{introduction_1}.

Despite high predictive accuracy, most current recommendation system methods rarely reveal explanatory behavioral patterns. 
Explainable recommendations have been proposed, aiming to accompany each recommendation with a human-interpretable explanation~\cite{explainablesurvey}. 
Early solutions~\cite{nrt, attn2seq} learned discrete ID-based embeddings for known users and items; however, this severely restricts their applicability to unseen users, who seldom possess extensive review histories and thus present a cold-start challenge. 
Recent work integrates advanced language models, including Transformer~\cite{erra}, GPT-2~\cite{yang2024fine}, and other large language models (LLMs)~\cite{kim2408review}, to leverage powerful generative and understanding capabilities.

\begin{figure}[htbp]
    \centering
    \small
    \includegraphics[width=0.87\columnwidth]{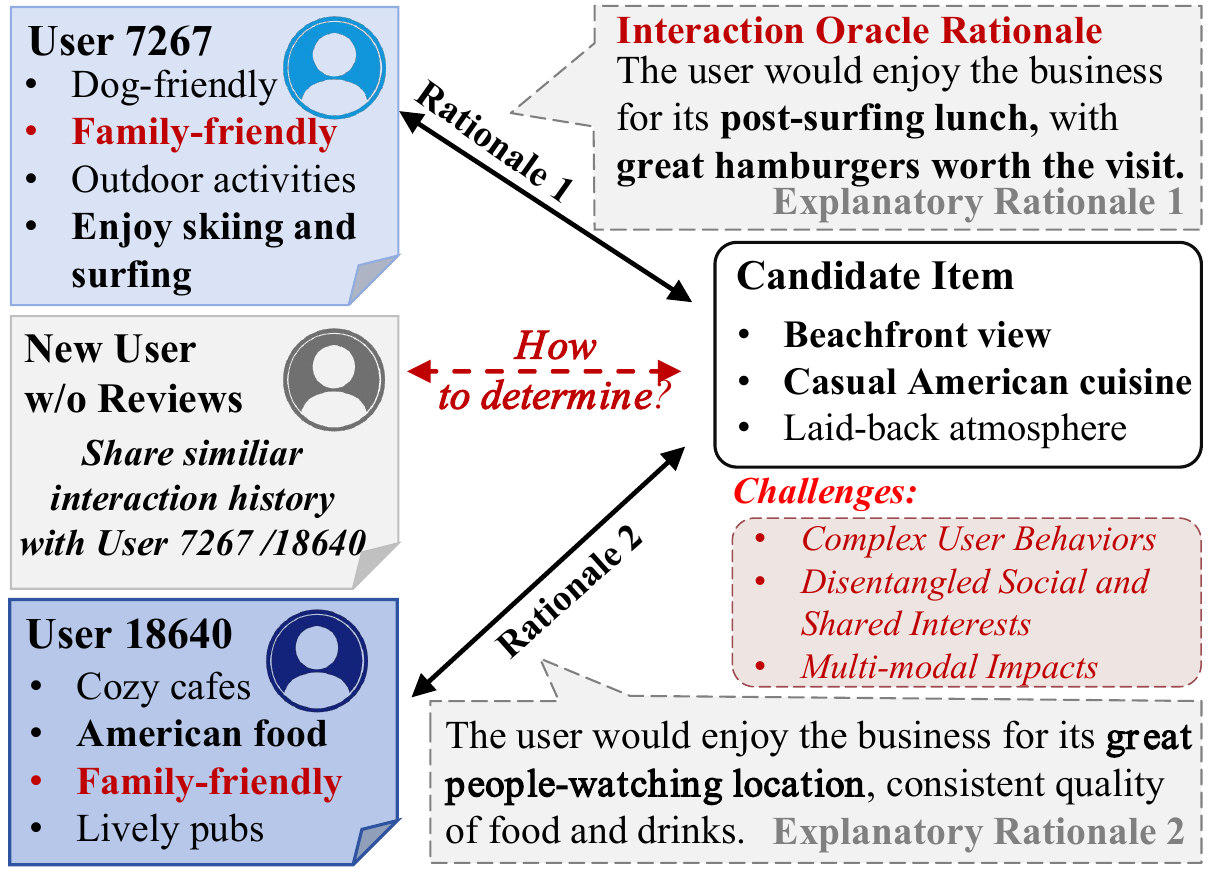}
    \caption{Illustration of challenges in real-world scenarios. Shared interests are highlighted in red, while the explanatory behavioral patterns are presented in bold.}
    \label{teaser}

\end{figure}

Existing approaches, however, still face three primary challenges as illustrated in Figure~\ref{teaser}.
First, their reliance on ID-level representations~\cite{peter} impedes generalization to cold-start users and items.
While graph-based methods offer a partial remedy~\cite{xrec}, they often suffer from the over-smoothing effect, which diminishes personalization~\cite{oversmoothing}.
For instance, users with similar interaction histories (e.g., user 7267 and user 18640) may engage with the same item for entirely different reasons. 
Thus, effectively representing users and items while balancing their collective and individual characteristics remains a significant difficulty. 
Second, many frameworks incur substantial computational overhead, as they either incorporate extensive profile texts into prompts or require fine-tuning LLMs~\cite{adaptingtokenllm}.
A core challenge is failing to efficiently distill explanatory behavioral patterns behind the interaction, which is fundamentally driven by user interests and item characteristics.
Third, most methods focus on either interaction history or review semantics in isolation, overlooking the valuable correlations between the two. 
Effectively unifying this heterogeneous information presents a persistent challenge.



Tackling these limitations can provide new insights into the design of explainable recommendation models.
In this paper, we introduce \textbf{\model}, a lightweight \textbf{Be}havior-\textbf{A}ware \textbf{T}okenizer that empowers a frozen LLM to reason about user-item interactions, eliminating the need for costly fine-tuning.
Specifically, review-enriched users/items are mapped to discrete \texttt{words} drawn from a shared \texttt{Behavior Vocabulary}, which captures collective trends while preserving fine-grained individual intents.
This condensed representation allows for harnessing the model’s reasoning power without the heavy computational overhead of lengthy textual profiles.
We feed these behavior tokens into the LLM, enabling it to use its built-in language knowledge to link user and item features and explain why each recommendation is made.
The vocabulary for these tokens is further refined using pre-trained textual correlations as supervision, thereby bridging collaborative filtering signals with natural language semantics.
Finally, we introduce an auxiliary alignment objective that embeds these behavior tokens directly into the LLM’s semantic space, enabling seamless integration.
To evaluate the effectiveness of the~\model, we conduct extensive experiments on three real-world datasets in zero-shot settings, which is more common in practical applications.
The results demonstrate that our approach effectively aligns linguistic and collaborative semantics through macro- and micro-level language supervision.  
Moreover, the proposed auxiliary alignment task further improves the LLM's ability to interpret behavior tokens within a unified semantic space.
The key contributions of this work are summarized:

\begin{itemize}
    \item We propose to learn a~\texttt{behavior vocabulary} that represents diverse users and items by capturing their mutual interests while preserving individual-level distinctions.  
    The vocabulary is developed through a multi-level cross-modality learning framework, leveraging both collaborative and textual semantic signals into supervision.
    \item Our new training paradigm focuses on lightweight, explainable recommendations. 
    It avoids computationally intensive processes and auxiliary model architectures.
    \item Extensive experiments validate the effectiveness of our approach.  
    Across-LLM evaluations confirm the tokenizer's transferability, while other analyses reveal how LLMs comprehend behavior tokens.
    Our behavior tokens effectively capture shared and distinct behavior patterns, transforming them into rich semantic representations. 
\end{itemize}







\section{Related Work}
\subsection{Explaniable Recommendation}
Early neural approaches generate review-style explanations based on explicit attribute inputs.
Att2Seq~\cite{attn2seq} encodes user–item pairs into an RNN decoder but relies on fixed attribute schemas.
NRT~\cite{nrt} jointly models rating prediction and explanation generation via shared latent factors, yet still depends on per-entity embeddings and historical ratings.
PEPLER~\cite{pleter} and PETER~\cite{peter} enhance interpretability by applying phrase-level attention to generate aspect-aware explanations without large-scale pretraining.
With the rise of LLM, XRec~\cite{xrec} further advances this paradigm by injecting collaborative filter signals into the embedding space of LLM, establishing a structured framework.
Recently, models such as Review-LLM~\cite{reviewllm}, Reason4-Rec~\cite{fang2025reason4rec}, and EXP3RT~\cite{kim2408review} rely on the retrieval or construction of user profiles.
However, this approach is computationally intensive, making it impractical for review-scarce scenarios. 
Moreover, these methods are constrained by their reliance on collaborative filtering for attribute extraction and architectural modifications to the LLM.
This thereby hinders their adaptability across evolving model backbones.

\subsection{Disentangled Recommendation Systems}
Disentangled representations of users’ latent intents have proven effective in refining preferences and improving the precision of recommendations~\cite{disenbuddle}.
MacridVAE~\cite{disengcn} models multiple user intents using a variational autoencoder in a structured latent space.
DGCF~\cite{dgcf} and DisenHAN~\cite{disenhan} employ graph-based methods with GNNs and attention mechanisms to disentangle interests.
Recent work~\cite{iclrec} integrates contrastive learning to enhance intent separation and representation discriminability.
The efficacy of existing disentangled intent models in explainable recommendation is constrained by their coarse-level preference modeling.
Such limitations hinder their applicability in explainable recommendations, where interpretability and semantic alignment are crucial for decision-making.
Our method encodes multi-modal information into a hierarchical structure of discrete tokens, capturing macro-level shared interests and micro-level individual preferences.
This architecture provides LLMs with the nuanced representations required for reasoning in complex scenarios.

\section{Problem Definition}
We aim to encode a user/item into $K$ informative discrete tokens, originating from a unified \texttt{behavior vocabulary} $\mathcal{B}$.
Although the user and item vocabularies are distinct in practice, we adopt a unified notation for simplicity.
Given an LLM vocabulary $\mathcal{T}$ consisting of $N$ language tokens, our objective is to establish a semantic alignment between $\mathcal{T}$ and the behavior vocabulary $\mathcal{B}$, allowing the LLM to perform explainable recommendations analogous to language generation based on given representations, 
\begin{equation}
    \text{Explanation}(u, i) = \text{LLM}(\text{Prompt}, \mathcal{B}_u^{k}, \mathcal{B}_i^{j}),
\end{equation}
where $\mathcal{B}_u^{k}$ and $\mathcal{B}_i^{j}$ denote the top-$K$ behavior tokens selected for user $u$ and item $i$, respectively, and $k, j \in \{1, \dots, K\}$.
The explanation in the output text reveals the explanatory behavior patterns behind the user-item interactions.

\begin{figure*}[th]
    \centering
    \includegraphics[width=1\textwidth]{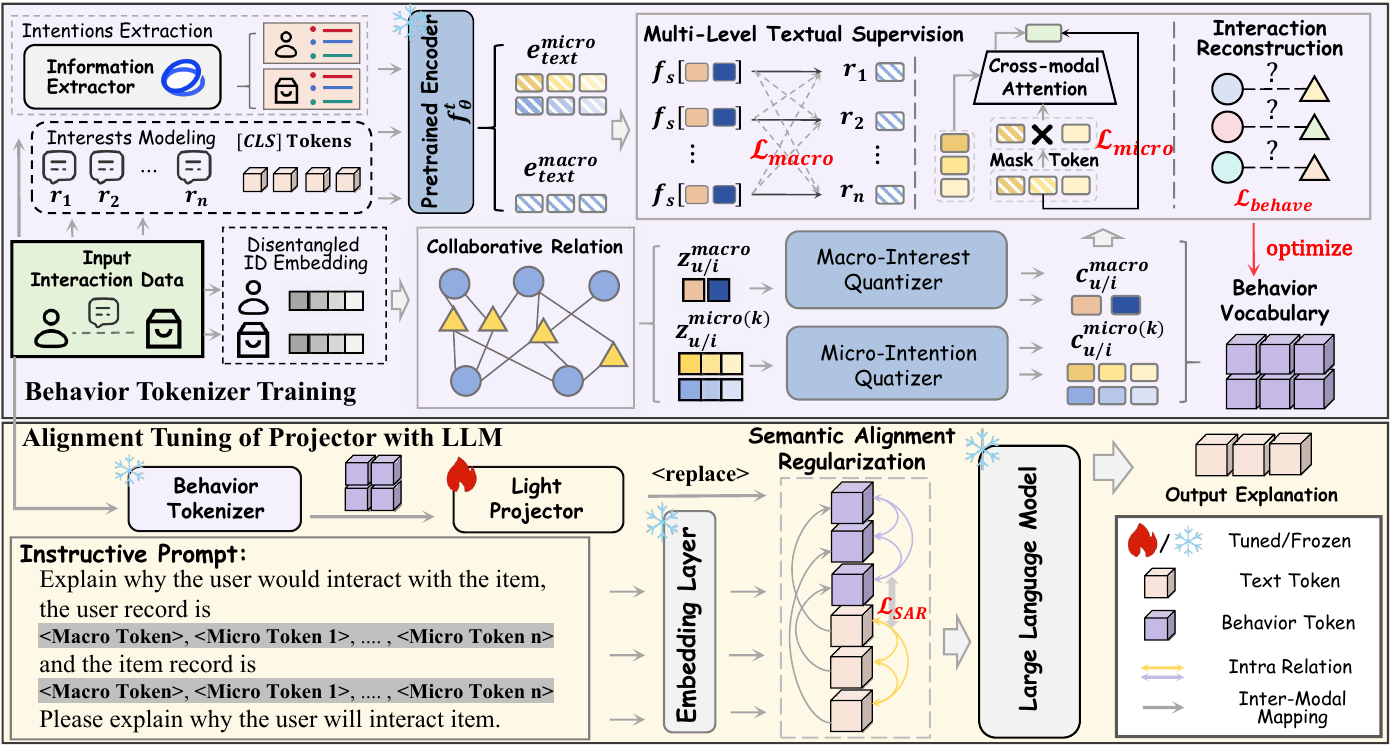}
    \caption{The \model~framework operates in two sequential stages. It employs a collaborative language tokenizer, supervised under multi-level language semantics, to encode diverse interactions into meaningful behavior tokens. Semantic alignment regularization is then introduced to refine the model by leveraging token-level correlations from the LLM's native vocabulary.}
    \vspace{-0.1in}
    \label{framework}
\end{figure*}
\section{Methodology}
As illustrated in Figure 1, our two-stage method first constructs a disentangled user representation, $\textbf{z}_u$, which captures a user's macro interests, $\textbf{z}^\text{macro}$, and micro intentions, $\textbf{z}^\text{micro}$. 
Following a graph propagation step, we select the most suitable macro and micro behavior tokens, $\textbf{c}^\text{macro}$ and $\textbf{c}^\text{micro}$, from the behavior vocabulary to align with the user's representation. 
We then employ multi-level semantic supervision with explicit language signals $\textbf{e}^\text{micro}$ and $\textbf{e}^\text{macro}$ to enhance the precision of our user modeling. 
Finally, a multi-objective training strategy, which combines semantic alignment regularization with next-token prediction, is utilized to establish a semantic correlation between the learned behavior vocabularies and the LLM's semantic space. An analogous procedure is followed for items.
\subsection{Disentangled Behavior Modeling}
A key challenge in enhancing LLMs for recommendations is effectively representing users and items as tokens within the model framework~\cite{hua2023index}. 
Existing methods assign unique IDs to each user and item, resulting in redundant vocabularies and the out-of-vocabulary (OOV) problem when adapting to new entities~\cite{adaptingtokenllm}. 
However, while users exhibit diverse preferences, they also share collective patterns shaped by trends or social factors. 
For instance, although preferences differ across product categories, common attributes such as affordability and brand reputation consistently remain priorities. 
To capture these collective and distinctive behaviors, we represent each user as a \texttt{sentence} composed of multiple \texttt{words}, where each word reflects an aspect of the user's preferences. 
The core idea is that similar users and items tend to share overlapping semantic tokens, allowing each unique token combination to correspond to a specific latent semantic pattern.


Formally, we aim to learn a factorized representation for each user \( u \), consisting of macro interests and micro intentions: \(\textbf{z}_u = [\textbf{z}_{u}^{\text{macro}} \,\|\, \textbf{z}_{u}^{\text{micro}(1)} \,\|\, \dots \,\|\, \textbf{z}_{u}^{\text{micro}(N)}] \in \mathbb{R}^{d'}\), where \( d' = (N + 1)d \). 
Here, \(\textbf{z}_{u}^{\text{macro}}\) captures user-specific macro-level preferences, while \(\textbf{z}_{u}^{\text{micro}(i)}\) represents the shared micro-level intention across users.
The operator \(\| \) denotes the concatenation of these components.
To incorporate collaborative relations among users, we adopt a lightweight graph convolutional approach~\cite{lightgcn} that propagates information through the user–item interaction graph:
\begin{equation}
    \mathbf{z}_u^{(l+1)} = \sum_{i \in \mathcal{N}_u} \frac{1}{\sqrt{|\mathcal{N}_u|} \sqrt{|\mathcal{N}_i|}} \mathbf{z}_u^{(l)}.
\end{equation}
After deriving collaboration-enhanced representations, we apply an averaging operation across graph propagation layers to preserve multi-order collaborative information:
\begin{equation}
    \mathbf{z}_u = \sum_{k=0}^K \frac{1}{K+1} \mathbf{z}_u^{(k)}.
\end{equation}

To capture the collectiveness and individuality of user behaviors, 
we train a Vector Quantized Variational Autoencoder (VQ-VAE)~\cite{vqvae} to discretize continuous input vectors by mapping them to a set of codewords.
For each user representation $\mathbf{z}_u$, we first decompose it into a set of interest- and intention-level vectors: $\{\mathbf{z}_u^{\text{macro}}, \mathbf{z}_u^{\text{micro}(i)}\}_{i=1}^N$.  
We then construct two separate codebooks of fixed size for macro-level interests and micro-level intentions.  
Together, these codebooks constitute the \texttt{behavior vocabulary} in our model, representing users in a discrete, interpretable token space, with codewords denoting \texttt{words}.
Taking the $i$-th micro-intention representation $\mathbf{z}_u^{\text{micro}(i)}$ as an example, we first project it into the vocabulary space $\mathbb{R}^{d'}$ via a linear transformation.  
We then compute its distance to all words in the vocabulary $\mathcal{C} = \{ \mathbf{c}_k \}_{k=1}^K$, and assign it to the nearest:
\begin{equation}
\textbf{c}_j = \arg\min_k \left\| \mathbf{z}_u^{\text{micro}(i)} - \mathbf{c}_k \right\|_2^2,
\end{equation}
where $\textbf{c}_j$ is the selected word assigned to the $i$-th input representation $\mathbf{z}_u^{\text{micro}(i)}$, and $\mathbf{c}_k$ denotes the $k$-th word in the corresponding vocabulary.
We derive a quantized representation for each user as $\textbf{q}_u = [\textbf{c}_{u}^{\text{macro}} \,\|\, \textbf{c}_{u}^{\text{micro}(1)} \,\|\, \dots \,\|\, \textbf{c}_{u}^{\text{micro}(N)}] \in \mathbb{R}^{M'}$, where $M' = (N + 1)M$ and $M$ denotes the codebook embedding dimension. 
The overall objective function for predicting user-item interaction using the quantized representations $\textbf{q}_u$ and $\textbf{q}_i$ is defined as follows:
\begin{equation}
\begin{aligned}
    &\mathcal{L}_{\text{RECON}} = \sqrt{ \left\| \mathbf{I}_{ui} - \mathbf{q}_u^\top \mathbf{q}_i \right\|_2^2 }, \\
    &\mathcal{L}_{\text{VQ-VAE}} = \left\| \text{sg}[\bm{q}] - \bm{c} \right\|_2^2 
    + \eta \left\| \bm{q} - \text{sg}[\bm{c}] \right\|_2^2, \\
    &\mathcal{L}_{\text{behave}} = \mathcal{L}_{\text{RECON}} + \mathcal{L}_{\text{VQ-VAE}},
    \label{eq:rqloss}
\end{aligned}
\end{equation}
where $\mathbf{I}_{ui} \in \{0, 1\}$ is a binary indicator of interaction between user $u$ and item $i$, $\text{sg}[\cdot]$ denotes the stop-gradient operator, and $\eta$ is a balancing hyperparameter, setting to 0.5.
The overall loss function comprises two components: a reconstruction loss $\mathcal{L}_{\text{RECON}}$ that encourages accurate prediction of user-item interactions, and a symmetric vector quantization loss $\mathcal{L}_{\text{VQ}}$ that minimizes the distance between the encoded representations and their assigned codebook embeddings.





\subsection{Multi-level Textual Semantic Supervision}
While Vector Quantization (VQ) excels at capturing common user behaviors~\cite{rajput2023recommender, lin2025unified}, its effectiveness is often hampered by codebooks constructed from unimodal data. 
Methods relying solely on collaborative signals face challenges with behavioral ambiguity, whereas those using textual data are constrained by its inherent sparsity.
These issues are compounded by data sparsity, as users in real-world systems typically interact with few items and provide limited explicit feedback~\cite{noiseinrecom}.
To overcome these issues, we propose a multi-level textual supervision approach to construct a unified codebook that integrates rich textual semantics with collaborative signals, thereby bridging the modality gap.

\noindent\textbf{Macro Semantic Supervision} aims to guide the learning of users’ macro-level interests using semantic signals derived from reviews.  
A user's macro interest is expected to be unique and distinguishable, as it reflects the fundamental and stable aspects of the user's overall preferences.  
Such macro-level interests may involve high-level goals, such as purchasing for broad categories of interest.

Given a review $r_{ui}$ associated with user $u$ and item $i$, we utilize a frozen, pretrained text encoder $f_{\theta}^{t}$ to extract its semantic representation. 
Considering that real-world reviews often vary in length and may contain irrelevant information, we avoid naively averaging all token embeddings. 
Instead, we adopt the \texttt{[CLS]} token as a global contextual feature that summarizes the review.
Specifically, we sample a batch of user–item pairs $\mathcal{B}$ along with their corresponding reviews, then apply the text encoder $f_\theta^t$ to obtain the semantic embeddings $\textbf{e}^{\texttt{[CLS]}}$, which serve as supervision signals for learning the macro behaviors.
Since a review reflects the interaction rationale between a user and an item, we introduce a fusion function $f_{s}$ to approximate this process as $\textbf{c}^{\text{macro}} = f_{s}(\textbf{c}_u^{\text{macro}}, \textbf{c}_i^{\text{macro}})$, where $\textbf{c}_u^{\text{macro}}$ and $\textbf{c}_i^{\text{macro}}$ are the quantized representations.
The reviewed user–item pairs are treated as positive instances, while all other pairs in the batch are considered negatives. 
Then the InfoNCE~\cite{infonce} loss is employed for bridging the semantic gap between reviews and behavior tokens, encouraging alignment between the review representation and the corresponding behavior token representations while distinguishing them from others, $s(\cdot, \cdot)$ is cosine similarity:

\begin{equation}
\scalebox{1}{$
\mathcal{L}_\text{macro}=- \sum_{i \in \mathcal{B}} \log \frac{
\exp \left( s(\textbf{e}^{\text{CLS}}_i, \textbf{c}^{\text{macro}}_i) \right)}{
\sum_{j \in \mathcal{B}} \exp \left( s(\textbf{e}^{\text{CLS}}_i, \textbf{c}^{\text{macro}}_j) \right)}.
$}
\end{equation}

\noindent\textbf{Micro Semantic Supervision} uses semantic cues from behavioral data to capture fine-grained user intentions.
This approach complements macro-level supervision, which identifies broad preferences that are often unique to an individual. 
In contrast, micro-intentions are frequently shared across many users—for instance, a common preference for features like "durability" or "ease of use," regardless of the product category.
Modeling users' micro-intentions toward specific items is achieved by extracting fine-grained intent expressions from their historical reviews. 
Given a user's historical review list, denoted as $\mathbf{R}_u = \{r_1, r_2, \cdots, r_n\}$, we prompt an LLM to perform an information extraction task over the review list.
This process retrieves a set of interpretable micro-intentions:
\begin{equation}\label{prompt}
\textbf{e}_{1}^{\text{micro}}, \cdots, \textbf{e}_n^{\text{micro}} = f_\theta(\text{LLM}(\mathbb{P}_{e}, \mathbf{R}_u)),
\end{equation}
where $\mathbb{P}{e}$ denotes the designed extraction prompt, and $\textbf{e}_{i}^{\text{micro}}$ is the $i$-th extracted micro-intention, encoded using a pretrained text encoder $f_\theta^t$ by preserving \texttt{[CLS]} tokens.

Despite we obtain both the textual micro-intention embeddings $\{\mathbf{e}_i^{\text{micro}}\}$ and the collaborative behavior token representations $\{\mathbf{c}_u^{\text{micro}(i)}\}$, establishing a direct one-to-one correspondence between them is intractable due to their latent and unordered characteristics.
To address this, we propose a fine-grained masked reconstruction strategy that leverages the semantic relationships among micro-intentions for supervision.
For every sequence of $n$ textual micro-intention representations, we randomly sample $t$ positions to be masked, resulting in a corrupted sequence  
$S^{\text{msk}} = \{\mathbf{e}_1^{\text{micro}}, \mathbf{m}_1, \mathbf{e}_3^{\text{micro}}, \cdots, \mathbf{e}_n^{\text{micro}}\}$,  
where $\mathbf{m}_i$ denotes a mask token at the $i$-th position.
We then employ a cross-attention module $f^c_\theta$ to incorporate the behavior token representations $\mathbf{c}^{\text{micro}}$ and infer the embeddings of the masked micro-intentions based on the contextual cues from the unmasked ones.  
The objective is to reconstruct the original semantic embedding of each masked intention:
\begin{equation}
    \mathcal{L}_\text{micro} = - \mathbb{E}_{(c_\text{micro}, S^\text{msk}) \sim \mathcal{B}}~\phi(e_{msk}, f^c_\theta (c_\text{micro}, S^\text{msk})),
\label{eq:mskloss}
\end{equation}
where $e_{msk}$ denotes the ground-truth embedding of the masked micro-intentions, and $\phi$ distance measuring function between distributions, which is the L2 distance in practice.
\paragraph{Overall Training Objective}  
The behavior vocabulary is optimized by integrating CF signals with textual semantic supervision, and is balanced by $\alpha$ and $\beta$:
\begin{equation}
    \mathcal{L}_\text{tokenizer} = \alpha \cdot \mathcal{L}_\text{macro} + \beta \cdot \mathcal{L}_\text{micro} + \mathcal{L}_\text{behave}.
\end{equation}

\subsection{Behavior Tokens Comprehension}
Multi-level semantic supervision establishes semantic correlations among the learned behavior tokens; however, a distributional gap persists between the LLM and the behavior tokens. 
We first introduce a natural language task prompt shown in Figure~\ref{framework} that contextualizes the behavior tokens.
This prompt design allows the LLMs to interpret the tokens in a semantically meaningful context, leveraging the behavior patterns and their intrinsic language understanding~\cite{liu2023visual}.
Previous methods typically tune LLMs to understand extensive tokens or explicitly modify LLM structures~\cite{adaptingtokenllm, xrec}, which limits their generalization in practical scenarios. 
We address this by introducing a projection module that maps behavior tokens directly into the LLM's semantic space with multiple training objectives. 
The projected token embeddings are then used to replace the placeholders $\mathsf{\langle Tokens \rangle}$, enabling the LLM to process the behavior tokens directly.


Since LLMs already possess rich semantic correlations among their native token spaces from massive pre-training, we propose to transfer these inherent semantic relationships to the newly introduced behavior tokens. 
For example, if a user is a history enthusiast, their behavior tokens and the behavior tokens of historical books should be associated with the ``history" aspect. 
Furthermore, the relationship between these specific tokens should reflect the textual semantic relationship between phrases like `love' and `historical books', thus catering to the LLM's intrinsic understanding.
We then propose \textbf{Semantic Alignment Regularization} as an auxiliary training objective to align the projected behavior tokens with the LLM’s native semantic space.

For every word representation $\epsilon_{w_i}$ encoded by the LLM from the explanation text, we aim to establish a semantic correlation between text attributes and behavior tokens. 
Since we have applied multi-level textual supervision, we map each text token to its corresponding behavior token:
\begin{equation}
    \hat{c}_i = \arg\min_{c \in \mathcal{C}} \left\| \epsilon_{w_i} - \mathbf{c} \right\|_2, 
\end{equation}
where $\mathcal{C}$ denotes the set of projected behavior token embeddings.
We then introduce the following loss function:
\begin{equation}
\mathcal{L}_{\text{SAR}} = \sum_{(w_i, w_j) \in \mathcal{T}} \left(s(\hat{c}_i, \hat{c}_j) - s(\epsilon_{w_i}, \epsilon_{w_j})\right)^2,
\end{equation}
where $\mathcal{T}$ is a sampled set of word pairs from the explanation text, and $s(\cdot, \cdot)$ denotes cosine similarity.  
This loss aligns the textual and behavioral semantic spaces by encouraging behavior tokens to reflect the semantic relationships found in the text. 
To enhance the coherence of the generated explanations, we train the model by minimizing the negative log-likelihood (NLL) of the ground-truth explanation sequence:
\begin{equation}
\mathcal{L}_{\text{NLL}} = -\frac{1}{N} \sum_{i=1}^{N} \sum_{c=1}^{C_i} y_{ic} \cdot \log(\hat{y}_{ic}),
\end{equation}
where $N$ denotes the number of explanations, $C_i$ is token number in the $i$-th explanation, and $y_{ic}$ and $\hat{y}_{ic}$ denote the ground truth and predicted probabilities for the $c$-th token.

\section{Experiments}
In this section, we conduct comprehensive experiments to evaluate the effectiveness of \textbf{\model}.  
These experiments are designed to address the following research questions:

\begin{itemize}
\item \textbf{RQ1:} How effective is \model~at generating explainable recommendations in a zero-shot setting, and what are the respective contributions of its core components?
\item \textbf{RQ2:} How robust and transferable is the proposed behavior tokenizer across different LLM backbones?
\item \textbf{RQ3:} How effectively does the behavior tokenizer capture user behavior patterns, particularly in cold-start scenarios involving unseen users and items?
\item \textbf{RQ4:} How does the model interpret the learned behavior tokens, and what semantic meaning do they convey?
\end{itemize}

\subsection{Experiment Settings}
The complete experimental settings are detailed in the Appendix. 
For all subsequent experiments, we define the `zero-shot' setting as users or items that have interaction data but lack any explicit textual reviews.




\subsection{Performance Comparison~(RQ1)}

\begin{table*}[t] 
    \centering
    \Large
    \resizebox{\textwidth}{!}{%
    \begin{tabular}{cc|ccc|ccc|ccc}
    \toprule
    \multirow{2}{*}{\textbf{Model type}} & \multirow{2}{*}{\textbf{Model}} & \multicolumn{3}{c|}{\textbf{Amazon}} & \multicolumn{3}{c|}{\textbf{Google}} & \multicolumn{3}{c}{\textbf{Yelp}} \\
    \cmidrule{3-11}
    & & \textbf{BLEU}$\uparrow$ & \textbf{BART}$\uparrow$ & \textbf{BERT}$\uparrow$ & \textbf{BLEU}$\uparrow$ & \textbf{BART}$\uparrow$ & \textbf{BERT}$\uparrow$ & \textbf{BLEU}$\uparrow$ & \textbf{BART}$\uparrow$ & \textbf{BERT}$\uparrow$ \\
    \midrule
    \multirow{3}{*}{\shortstack{ID Based}} 
    & Attn2seq & 0.2843 & -4.9738 & -0.1207 & 0.3846  & -4.2112 & 0.3101 & 0.3834 & -4.6040 & 0.1841 \\
    & NRT & 0.2366 & -4.3370 & 0.3300 & 0.3045  & -4.1964 & 0.2965 & 0.3112 & -5.0841 & 0.2652 \\
    & PETER & 0.3682 & -4.2300  & 0.1488 & 0.3533 & \textbf{-3.6307} & 0.3283 &          0.3329           &   -4.6469           & 0.1675         \\
    \midrule
    \multirow{6}{*}{\shortstack{LLM Based}} 
    & PLETER & 0.3120  & -4.0630  & 0.2816 &      0.3850               &        \underline{-4.1243}      &   0.3631       &   0.3470                  &   -4.5996           &  0.3252        \\
    & XRec & 0.2999  & -4.3210 & \underline{0.3598} &     0.2785                &    -4.5901          & 0.2628         & 0.3565  & -4.6226 & 0.2871 \\
    & XRec-Profile & 0.3854 & -4.1749 & 0.3259 & 0.3289  & -4.3491 & 0.3251 & 0.2925  & -4.5955  & 0.2579 \\
    & GraphGPT & 0.3596  & -4.0380 & 0.3494 & 0.3702 & -4.4328 & 0.3650 & 0.3399 & -4.5275  & 0.2958 \\
    & TEA-GLM & \underline{0.3971}  & -4.1348 & 0.3406 & 0.3689 & -4.3574 & 0.3521 & \underline{0.3844}  & -4.5452 & 0.3067 \\
    & Time-LLM & 0.3936  & -4.0780 & 0.3470    & 0.2727 & -4.2873 & 0.2983 & 0.3368  & -4.5933 & 0.3135 \\
    \midrule
    \multirow{4}{*}{\shortstack{Our Proposed}} 
    & \textbf{\model~w/o Micro} & 0.3781 & \underline{-4.0600} & 0.3447 & 0.3027  & -4.2686 & 0.3397 & 0.3136  & -4.5545 & 0.2793 \\
    & \textbf{\model~w/o Macro} & 0.3795  & -4.1382 & 0.3415 & 0.3834  & -4.3384  & 0.3580 & \textbf{0.3852} & -4.5691 & 0.3272  \\
    & \textbf{\model~w/o SAR} & 0.3720  & -4.0945 & 0.3531 & \textbf{0.3922}  & -4.3939  & 0.3776 & 0.2967  & \textbf{-4.5019} & 0.3124 \\
    & \textbf{\model}  & \textbf{0.4195}  & \textbf{-3.9929} & \textbf{0.3821} & \underline{0.3866} & -4.3027 & \textbf{0.3781} & 0.3771 & \underline{-4.5442} & \textbf{0.3302} \\
    \bottomrule
    \end{tabular}
    }
    \caption{Zero-shot accuracy on Amazon, Google, and Yelp datasets. Each dataset's evaluation metrics include BLEU, BARTScore, and BERTScore. (\textbf{bold} highlights the best result across all methods, while \underline{underline} highlights the second)}
    \vspace{-0.08in}
    \label{table:all}
\end{table*}
Experiments conducted on three real-world datasets demonstrate the effectiveness of our proposed \model. 
We choose three different metrics to validate the performances, as BERTScore~\cite{bertscore}, BARTScore~\cite{bartscore}, and BLEU~\cite{bleu}.
The results, presented in Table~\ref{table:all}, show that \model~achieves state-of-the-art or highly competitive performance across all three scenarios.
Traditional ID-based methods, such as Attn2Seq or NRT, appear to struggle in zero-shot settings, indicating the limited expressiveness of ID representations.
While LLM-based methods have usually achieved promising results, demonstrating the effectiveness of LLMs' generation ability.
Recently popular cross-modality alignment methods also show competitive results, showing the necessity of semantic alignment; our model effectively integrates these strengths.
A detailed ablation study isolates the contributions of our model's key components: the micro tokens, macro tokens, and the second-stage alignment. 
Removing the micro tokens led to a significant performance drop, confirming that fine-grained representations for users and items are essential. 
The impact of removing macro tokens, however, was dataset-dependent: it degraded performance on Google but improved it on Amazon and Yelp. 
We conjecture that a high-level summary token may occasionally distract the LLM from detailed user preferences. 
Disabling the second-stage alignment caused a substantial performance decrease, particularly on the complex Yelp dataset. 
This result underscores that for sophisticated domains, incorporating semantic relationships through alignment is necessary to grasp the underlying recommendation rationale fully.

\subsection{Robustness across LLM Backbones~(RQ2)}
\begin{figure}[h]
    \centering
    \small
    \includegraphics[width=0.8\columnwidth]{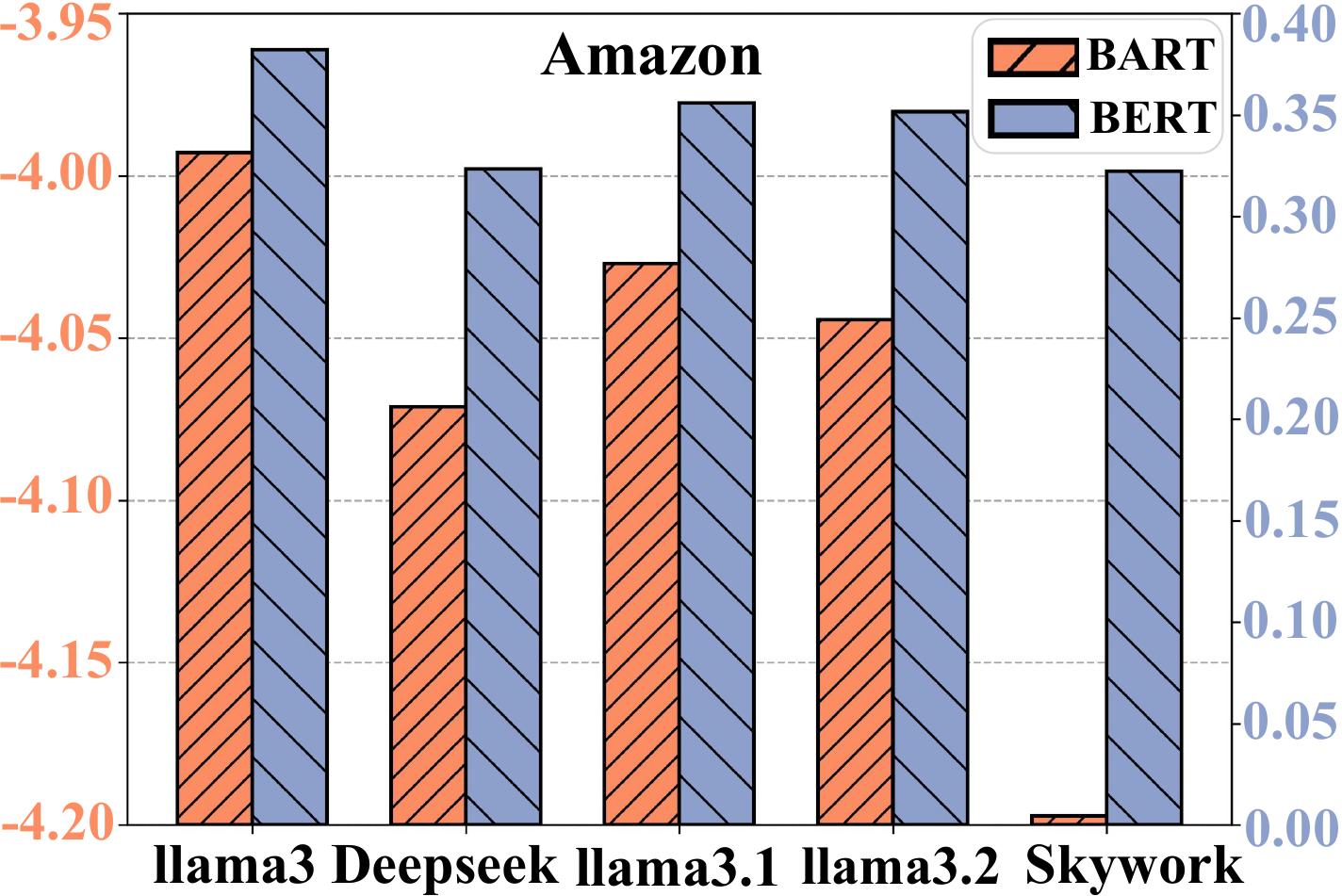}
    \caption{Robustness of~\model~Across LLM Backbones}
    \label{across_backbones}
\end{figure}

To assess the robustness and transferability of \model, we evaluated its performance with several open-source LLM backbones: DeepSeek-8B~\cite{deepseek}, LLaMA3.1-8B, LLaMA3.2-3B~\cite{llama_models}, and Skywork-8B~\cite{skywork}.
The evaluation was conducted on the Amazon dataset, with results reported in Figure~\ref{across_backbones} using BARTScore and BERTScore-F1 as metrics.
The LLaMA models (LLaMA3.1-8B and LLaMA3.2-3B) achieved performance comparable to LLaMA3-8B, suggesting that models with similar architectures share compatible semantic spaces. 
Notably, the much smaller LLaMA3.2-3B delivered strong results, demonstrating our method's scalability and effectiveness in resource-constrained settings.
In contrast, DeepSeek-8B and Skywork-8B showed lower fluency and fidelity, which we attribute to their known tendency for hallucination. 
While these backbones are less competitive, they still surpassed most baselines.
Overall, these findings confirm the robustness of \model~across various backbones and its straightforward adaptability to suit practical needs.

\begin{figure}[h]
\center
\includegraphics[width=0.97\columnwidth]{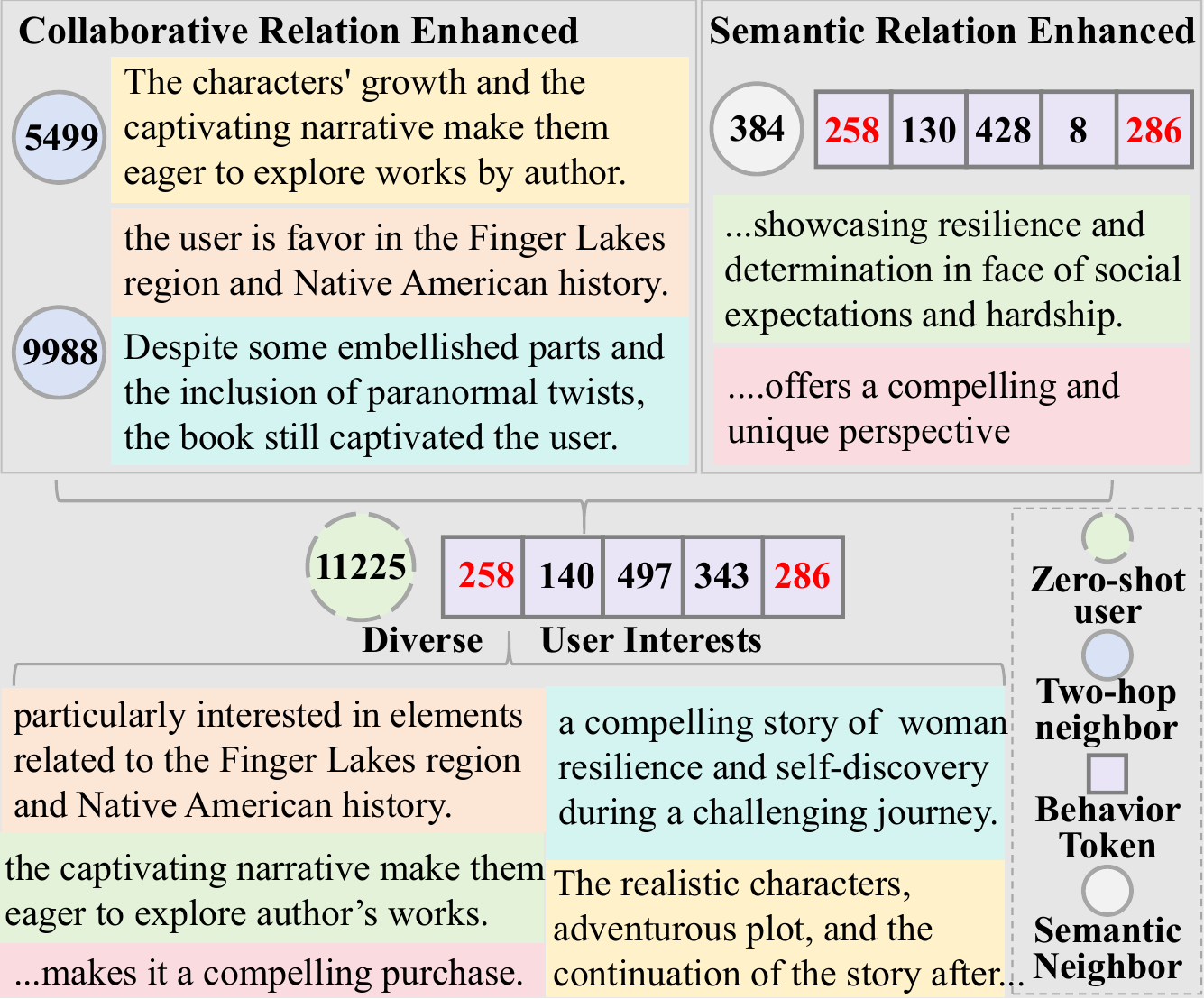}
\caption{A Token-Level Visualization for Zero-Shot User.}
\label{fig:casestudy}
\end{figure}
\begin{figure*}[!t]
\center
\includegraphics[width=0.95\textwidth]{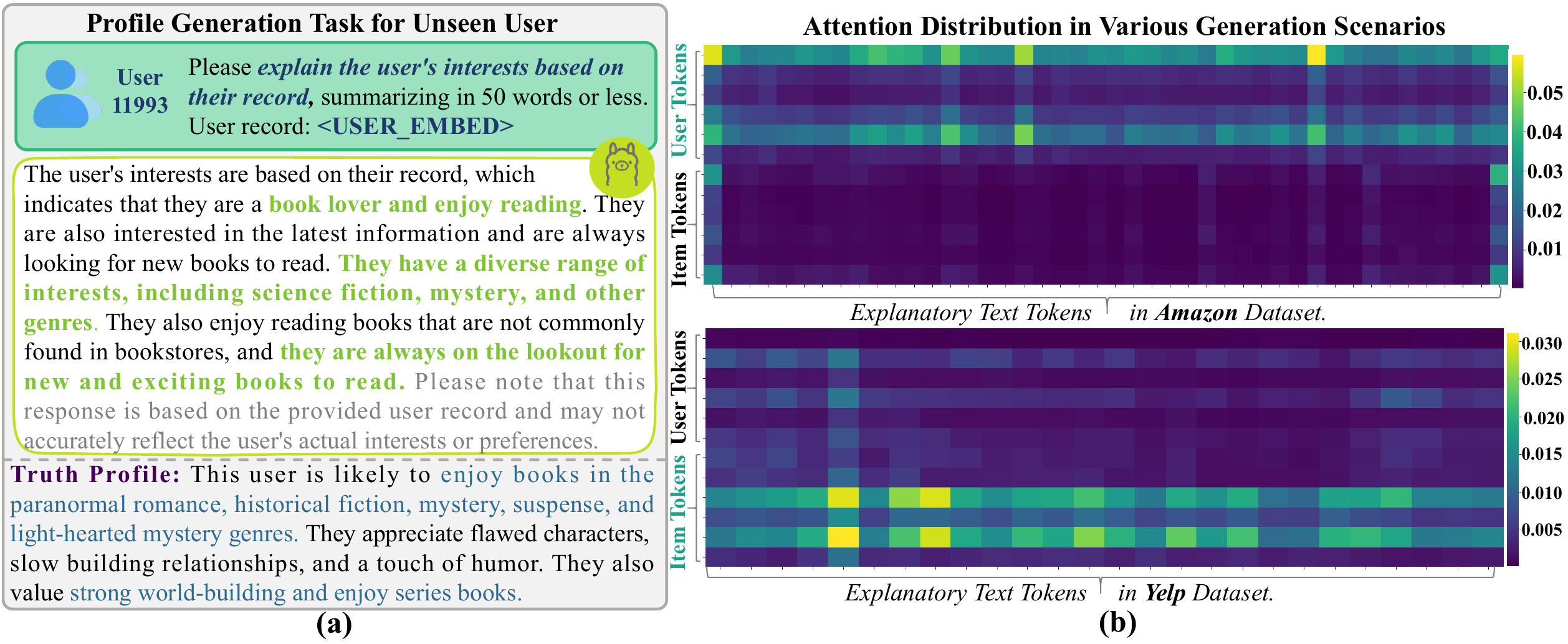}
\caption{Illustration of the LLM's Comprehension of Behavior Tokens. (a)The LLM's generated response when conditioned on a batch of behavior tokens and an instruction prompt. (b)Visualization of the attention distribution over input behavior tokens.}
\label{fig:profile_generation}

\end{figure*}
\subsection{Modeling Behavior for Cold-Start Users~(RQ3)}

Figure~\ref{fig:casestudy} presents a case study on a ``cold-start" user to demonstrate how our model assembles a token representation for them without any review history.
The model first constructs an initial profile by borrowing from semantic neighbors; for instance, it uses tokens from User 384 to infer an interest in themes like "resilience and self-discovery" (Token 286).
This profile is then refined with collaborative signals from users with overlapping interests, such as User 5499 and User 9988, who both share an interest in the Finger Lakes region and Native American history.
As illustrated by the color-coding in the figure, this process weaves together semantic and collaborative information to bridge low-level symbolic tokens with high-level interests.
Ultimately, the model produces a coherent and interpretable token set for a completely new user, showcasing its robustness in overcoming the cold-start limitations of traditional methods.

\subsection{Interpretability of Behavior Tokens~(RQ4)}
\paragraph{User Profile Generation} Despite the effectiveness of the proposed behavior tokens, it remains unclear whether the LLM is simply memorizing input patterns or understanding their semantic correlations. 
To investigate this, we prompted the LLM to generate a user profile directly, without any task-specific training. 
Figure~\ref{fig:profile_generation}(a) presents the result for a zero-shot user, whose reviews are ablated from all training stages.
The results demonstrate that the LLM successfully inferred the user's preferences and interests.
Notably, it also correctly deduced the book-purchasing context without explicit cues, indicating that the scenario's semantics were properly encoded. 
The generated profile offered detailed explanations for the user's interests and largely matched the ground-truth. 
Although some hallucinations~\cite{yao2023llm} were observed, we posit that they can be mitigated through targeted fine-tuning. 
Given that the LLM was not trained on this task, these findings suggest that it has established a semantic correlation with the behavior tokens, rather than memorizing.

\paragraph{Behavior Token Functioning} Although our experiments demonstrate that the LLM can holistically understand behavior tokens, the underlying working mechanism remains obscure.
This prompted us to investigate the functional effectiveness of these individual behavior tokens. 
Figure~\ref{fig:profile_generation}(b) visualizes the attention matrices for two distinct scenarios: the Amazon dataset (top) and the Yelp dataset (bottom). 
In each matrix, the color intensity reflects the attention score between the input tokens on the left (partitioned into user- and item-specific sets) and the generated words on the bottom.
Specifically, the model's attentional focus shifts from users on the Amazon dataset to items on the Yelp dataset.
We conjecture that this is because the Amazon dataset covers a narrow domain: book reviews. 
In such a homogeneous context, the rationale for a user-item interaction is likely to depend on the user's specific interests. 
In contrast, the Yelp dataset encompasses a diverse range of items, the unique attributes of the item itself become more critical for generating a meaningful rationale.
This finding demonstrates the model's ability to dynamically shift its focus between users and items to suit the dataset's context. 
Furthermore, different tokens make distinct contributions during the generation process, revealing the complexity of user behavior.
Our model leverages these varied signals to craft explanations that reflect the nuanced interaction between a user's diverse preferences and an item's specific attributes.


\section{Conclusion}

In this work, we introduced a unified behavior tokenizer that translates user-item interactions into an LLM-comprehensible vocabulary. 
By leveraging a two-level textual supervision mechanism and incorporating semantic relational knowledge, our method enables LLMs to understand these behavioral tokens without extra fine-tuning, achieving state-of-the-art or highly competitive performance on benchmark datasets. 
Further analysis reveals how our approach captures diverse user interests, enhances the LLM's capacity for behavioral reasoning, and how the LLM interprets given behavior tokens.
Future work will focus on extending this approach to diverse recommendation scenarios and evaluating its cross-domain transferability.


\section{Acknowledgments}
This work was supported by National Key R\&D Program of China (2024YFB4506004), National Natural Science Foundation of China (62272023), and partly supported by Science and Technology Project of Beijing Municipal Commission of Transport (2025-KJC-03-003).

\bibliography{aaai2026}
\appendix
\section{Appendix}

\subsection{Experimental Settings}\label{experiment settings}
\paragraph{Datasets} We evaluate \textbf{\model} using three public datasets: \textbf{Amazon}~\cite{amazon}, \textbf{Yelp}~\cite{yelp}, and \textbf{Google}~\cite{google}.
These datasets, sourced from~\cite{xrec}, each offer a distinct perspective on user-item interactions.
Acknowledging that reviews are often scarce while interactions are abundant in real-world scenarios, our evaluation specifically focuses on zero-shot conditions. 
Table~\ref{table:dataset} provides detailed statistics for these datasets, where "\texttt{\#Interactions}" denotes non-review interactions between users and items, and "\texttt{\#Reviews}" refers to user-generated reviews for items.

\begin{table}[h]
    \centering
    \small
    \caption{Statistics of the experimental datasets.}
    \begin{tabular}{ccccc}
        \toprule
        Dataset & \#Users & \#Items & \#Reviews &\#Interactions\\
        \midrule
        Amazon & 15,349 & 15,247 &  31,849 & 360,839 \\
        Yelp & 15,942 & 14,085 & 52,085 &  393,680\\
        Google & 22,582 & 16,557 &  31,939 & 411,840 \\
        \bottomrule
    \end{tabular}
    \label{table:dataset}
\end{table}

\paragraph{Baseline Description} \label{baseline}
To comprehensively evaluate the performance of our method, we compare it against a diverse set of baselines, categorized into the following two groups:
i) Traditional Explainable Recommendation Models.
These methods rely on ID-based representations, including
\textbf{Att2Seq}~\cite{attn2seq}, \textbf{NRT}~\cite{nrt}, and \textbf{PETER}~\cite{peter}.
ii) LLM-Based Explainable Recommendation Models.
These approaches leverage pretrained large language models to enhance both recommendation quality and explanation generation, including \textbf{PLETER}~\cite{pleter}, \textbf{XRec}~\cite{xrec}, \textbf{GraphGPT}~\cite{graphgpt}, \textbf{TEA-GLM}~\cite{teaglm}, and \textbf{Time-LLM}~\cite{timellm}.
Notably, although GraphGPT, TEA-GLM, and Time-LLM are not specifically designed for explainable recommendation, they align graph-based structural signals with LLMs' semantic spaces through designed supervised learning.
Although there are also several baselines like \textbf{Review-LLM}~\cite{reviewllm}, \textbf{EXP3RT}~\cite{kim2408review}, and \textbf{PRAG}~\cite{prag}, they can be applied under our experimental settings.
\paragraph{Implementation Details} \label{implementation}
The experiments are run on an NVIDIA RTX 3090 with 24 GB RAM.
For the behavior tokenizer training, we configure 5 micro-tokens and 1 macro-token for each user and item. The codebook size is set to 512, with the loss parameters $\alpha$ and $\beta$ set to 0.2 and 1, respectively.
To align with the language model's requirements, the embedding size is set to 768, the learning rate is set to 1e-3, and for training efficiency, a batch size of 1024 is employed.
We apply an early stopping mechanism based on HitRatio@20, with a patience of up to 5 steps.
The SELFRec framework~\cite{yu2023self} is adopted for our training process.
In the second stage, we fine-tune the LLM, using LLaMA3-7B as our base model unless otherwise specified.
A two-layer MLP serves as a lightweight projector to align the behavior tokens with the LLM. 
To improve efficiency, we accelerate training using the DeepSpeed library with ZeRO Stage 2 optimization and 8-bit quantization.
The model is trained for 3 epochs with a batch size of 8, a learning rate of 1e-4, and a weight decay of 1e-6.
The loss weight $\gamma$ is set to 1.
All generated explanations are limited to a maximum of 50 words to ensure conciseness and clarity.
To adapt non-explainable models, such as GraphGPT, Time-LLM, and TEA-GLM, we utilized their built-in alignment mechanisms while incorporating our custom behavior tokens.

\paragraph{Evaluation Metrics} \label{evaluation}
When evaluating our \model, we employ a suite of metrics designed to capture semantic explainability from multiple perspectives, including BERTScore, BARTScore, and BLEU.
\textbf{BERTScore}~\cite{bertscore} measures token-level semantic similarity by computing the cosine similarity of contextualized embeddings from a pretrained BERT model.
\textbf{BARTScore}~\cite{bartscore} frames evaluation as a generation task, assigning scores based on the likelihood of the model reconstructing reference texts using the BART language model.
\textbf{BLEU}~\cite{bleu} is a traditional n-gram overlap metric that evaluates surface-level similarity between generated and reference texts.
The first two metrics assess the semantic alignment between the generated explanations and ground-truth references, focusing on meaning preservation rather than surface form.
To complement these, we include BLEU-1 as a precision-based lexical metric, allowing for direct comparison with prior text-generation baselines and highlighting surface-level fluency and consistency. 

\subsection{Prompt Example} 
In this section, we present the prompt template employed in Equation 7.
We selected Chat-GLM-Flash-4~\cite{glm2024chatglm} as our LLM-based information extractor due to its efficiency and cost-effectiveness.
The prompt is shown as:

\newpage
\begin{tcolorbox}[
    enhanced,
    colback=white,             
    colframe=black,            
    boxrule=0.8pt,             
    arc=3pt,                   
    colbacktitle=black,        
    coltitle=white,            
    fonttitle=\bfseries,       
    title=User Profile Extraction, 
    sharp corners=south,       
    left=2mm
]
\small 
\texttt{You are tasked with analyzing user profile descriptions to extract key user preferences and characteristics.}\\
\texttt{I will provide a series of user history reviews as a single string.}\\
\texttt{Your task is to identify and extract the user's preferences, traits, and interests in a comprehensive manner based on the input.}\\
\texttt{Please strictly follow these instructions:}\\
\begin{enumerate}
    \item \texttt{Extract as many meaningful preferences or characteristics as possible from the user's history reviews.}
    \item \texttt{Summarize each preference or trait clearly and concisely (a short phrase or single sentence).}
    \item \texttt{Rank the extracted preferences by their importance or prominence in the user's profile --- the most important should appear first.}
    \item \texttt{Output the result strictly in the following JSON format:}
\end{enumerate}
\begin{verbatim}
{
  "preferences": [
    "preference_1",
    "preference_2",
    ...,
    "preference_n"
  ]
}
\end{verbatim}
\texttt{Do not include any explanation, comment, or extra text --- only return the JSON object.}

\texttt{[...Detailed User Review Histories...]}\\
\end{tcolorbox}

\end{document}